\documentclass[sigconf,nonacm]{aamas}

\usepackage{balance} 

\doi{QNKH4630}

\usepackage{tikz}
\usepackage{xspace}
\usepackage{amsmath}
\usepackage{amsfonts}
\usepackage{mathdesign} 
\usepackage{upgreek}

\usepackage{cleveref}
\usepackage{paralist}

\title[Contextual Intelligence]{Contextual Intelligence\protect\\The Next Leap for Reinforcement Learning}
\subtitle{Blue Sky Ideas Track}

\author{\href{https://orcid.org/0000-0002-8703-8559}{André Biedenkapp}}
\affiliation{
 \institution{Albert-Ludwigs-Universität}
 \city{Freiburg}
 \country{Germany}}
\email{biedenka@cs.uni-freiburg.de}

\usepackage{blindtext}
\let\oldcite\cite

\renewcommand{\cite}[1]{%
 \def\tempcitekey{#1}%
 \def\todocitekey{TODO}%
 \ifx\tempcitekey\todocitekey
  {\color{orange}\bfseries\oldcite{#1}}%
 \else
  \oldcite{#1}%
 \fi
}

\begin{abstract}
Reinforcement learning (RL) has produced spectacular results in games, robotics, and continuous control. Yet, despite these successes, learned policies often fail to generalize beyond their training distribution, limiting real-world impact. Recent work on contextual RL (cRL) shows that exposing agents to environment characteristics -- \emph{contexts} -- can improve zero-shot transfer. So far, the community has treated context as a monolithic, static observable, an approach that constrains the generalization capabilities of RL agents.

To achieve contextual intelligence we first propose a novel taxonomy of contexts that separates \emph{allogenic} (environment-imposed) from \emph{autogenic} (agent-driven) factors. We identify three fundamental research directions that must be addressed to promote truly contextual intelligence:
\begin{inparaenum}
 \item \textbf{Learning with heterogeneous contexts} to explicitly exploit the taxonomy levels so agents can reason about their influence on the world and vice versa;
 \item \textbf{Multi-time-scale modeling} to recognize that allogenic variables evolve slowly or remain static, whereas autogenic variables may change within an episode, potentially requiring different learning mechanisms;
 \item \textbf{Integration of abstract, high-level contexts} to incorporate roles, resource \& regulatory regimes, uncertainties, and other non-physical descriptors that crucially influence behavior.
\end{inparaenum}

We envision context as a first-class modeling primitive, empowering agents to reason about \textit{who} they are, \textit{what} the world permits, and \textit{how} both evolve over time. By doing so, we aim to catalyze a new generation of context-aware agents that can be deployed safely and efficiently in the real world.
\end{abstract}


\newcommand{\BibTeX}{\rm B\kern-.05em{\sc i\kern-.025em b}\kern-.08em\TeX}

\newcommand{\MDP}[0]{M}            
\newcommand{\statespace}[0]{\mathcal{S}}        
\newcommand{\stateRL}[0]{s}                
\newcommand{\actionspace}[0]{\mathcal{A}}        
\newcommand{\transdomain}[0]{\mathcal{T}}        
\newcommand{\rewards}[0]{\mathcal{R}}          
\newcommand{\reward}[0]{r}               

\newcommand{\policy}[0]{\pi}              

\newcommand{\cMDP}[0]{\mathcal{M}}           

\newcommand{\context}{c}
\newcommand{\contextset}{\mathcal{C}}
\newcommand{\MDPc}{\MDP_\context}
\newcommand{\contextdist}{p_\contextset}

\newcommand{\CMDP}{cMDP\xspace}

\usetikzlibrary{shapes.geometric}
\usetikzlibrary{positioning,shapes,shadows,arrows,calc}
\usetikzlibrary{intersections}
\pgfdeclarelayer{background}
\pgfdeclarelayer{foreground}
\pgfsetlayers{background,main,foreground}

\tikzstyle{activity}=[rectangle, draw=black, rounded corners, text centered, text width=6em, fill=white, drop shadow]
\tikzstyle{data}=[rectangle, draw=black, text centered, fill=black!10, text width=6em, drop shadow]
\tikzstyle{myarrow}=[->, thick]
\usepackage{booktabs}
\usepackage{tabularx}

\begin{document}


\pagestyle{fancy}
\fancyhead{}


\maketitle 


\section{Introduction}
Reinforcement learning (RL) \cite{sutton-book18a} is a powerful paradigm that enables training of intelligent agents capable of solving even highly complex tasks.
The simplicity of this paradigm promises great flexibility and the potential to be applicable to a large variety of target domains.
However, prominent success stories of RL have largely focused on application domains with highly accurate, high-fidelity simulators. 
For example, the ready availability of game engines has spawned an increased interest in RL research, fueled by a string of impressive results from playing Atari Games \citep{mnih-nature15a} over mastering StarCraft II \cite{vinyals-nature19a} to generally capable racing policies \cite{wurman-nature22a,kaufmann-nature23a,sparc-aaai26a}.
Beyond game playing, RL has been used to learn policies for magnetic control of Tokamak plasmas \citep{degrave-nature22a} or navigation of stratospheric balloons with difficult to predict weather conditions \citep{bellemare-nature20a}.
Despite these impressive successes, we believe RL research is largely held back by relying on having access to, or being able to design ``perfect'' environments. 
Consequently, RL agents are typically not able to be transferred to settings that are even slightly different from their training environment \citep{zhang-iclr21a,benjamins-tmlr23a,kirk-jair23a,gumbsch2024learning,prasanna-rlc24a} or potentially need additional environment interactions at deployment to be able to act optimally in novel environments
\citep{offlineRLtut,beck-tmlr25a}.

A popular research direction in RL and robotics focused on exposing the learning agents to a wider distribution of experiences to mitigate this limitation. Domain randomization (DR) \citep{tobin-iros17a,peng-icra18a} and procedural content generation (PCG) \citep{cobbe-icml20a,wang-neurips21a} train agents on a distribution of related environments. This forces agents to learn robust behaviors rather than overfitting to particularities of a single environment. While these agents can be expected to be more transferrable across (highly) similar environments \citep{tobin-iros17a,andrychowicz-ijrr20a,klink-neurips20a}, these generalization capabilities often result in suboptimal solutions on individual environments \citep{reed2022a,prasanna-rlc24a}. This is to be expected however, as such agents are not explicitly aware about which environment they are acting on, and thus they need to learn behaviors that work well on average. In a similar vein, robustness can be achieved by modeling this objective as a min-max problem. In this setting the goal is to learn a policy that maximizes the reward under the worst possible adversarial setting \citep{robust_rl_adversary_neurips20,robust_rl_offline_neurips22}. This approach can mitigate worst-case outcomes but further sacrifices performance in average case scenarios as the learned policies act highly conservatively.

Contextual reinforcement learning (cRL) \citep{benjamins-tmlr23a} offers a more principled approach to generalization by making environment characteristics, the so called context \citep{hallak-arxiv15a,modi-alt18a}, explicit in the training of agents. Contexts could, for example be physical properties of the system and consist of masses of a robot \citep{prasanna-rlc24a} or payloads \citep{9294027}, surface conditions \citep{iannotta2025contextbridgerealitygap} and decision time \citep{biedenkapp-icml21a}. Such cRL works assume that context is either readily available, e.g., from sensor readings \citep{benjamins-tmlr23a,kirk-jair23a,prasanna-rlc24a,sparc-aaai26a}, or that it is unobservable and needs to be inferred \citep{yu2017preparing,reed2022a,ndir-ewrl24a,benad2025shared}. Learning in such a manner facilitates much improved zero-shot generalization capabilities \citep{kirk-jair23a} as agents can learn how to adapt to the environment such that they can act optimally on every environment and not just in the average case.

While cRL has proven to be effective in learning generalizable policies, especially with respect to zero-shot generalization, cRL still lacks a principled understanding and mechanisms to
\begin{inparaenum}
  \item \textit{learning from observations and potentially heterogeneous contexts};
  \item \textit{designing architectures and learning rules that directly leverage contextual structure};
  \item \textit{integrate abstract, high-level contexts such as uncertainty of observations, roles of agents in a multi-agent system or resource budgets}.
\end{inparaenum}
Solutions to these challenges will advance the field of multi-agent systems to unlocking truly contextual intelligence.
The rest of the paper elaborates the contextual RL problem, provides a novel taxonomy of contexts based on which we sketch potential solution approaches to the three challenges we identified.

\section{Learning Generalizable Policies}
\begin{figure}[tbp]
  \centering
  \begin{tikzpicture}[node distance=2.1cm]%
%
\node (Agent) [activity,
minimum height=.75cm] {Policy $\varpi$ or $\policy$};%
\node (Algo) [activity, right of=Agent, xshift=3cm,
minimum height=.75cm, text width=8em] {Environment $\MDPc$};%
\begin{pgfonlayer}{background}%
\path (Agent -| Agent.west)+(-0.12,1.25) node (resUL) {};%
\path (Algo.east |- Algo.south)+(0.125,-1.25) node (resBR) {};%
%
\path [rounded corners, draw=black!50, fill=white] ($(resUL)+(0.375, -0.375)$) rectangle ($(resBR)+(0.375, -0.375)$);%
\path [rounded corners, draw=black!50, fill=white] ($(resUL)+(0.25, -0.25)$) rectangle ($(resBR)+(0.25, -0.25)$);%
\path [rounded corners, draw=black!50, fill=white] ($(resUL)+(0.125, -0.125)$) rectangle ($(resBR)+(0.125, -0.125)$);%
%
\path [rounded corners, draw=black!50, fill=white] (resUL) rectangle (resBR);%
\path (resBR)+(-.9,0.175) node [text=black!75] {context $c$};%
\end{pgfonlayer}%
\draw[myarrow] (Agent.north) -- ($(Agent.north)+(0.0,+0.35)$) -- ($(Algo.north)+(0.0,+0.35)$) node [below=-.05,pos=0.5] { action $\policy(s_t) \to a_t$} node [above=-.05,pos=0.5] { action $\varpi(s_t,c) \to a_t$} -- (Algo.north);
\draw[myarrow, dashed] ($(Algo.south)+(-0.25, 0)$) -- ($(Algo.south)+(-0.25, -0.35)$) -- ($(Agent.south)+(0.25, -0.35)$) node [above,pos=0.5] {state $s_{t+1}$} -- ($(Agent.south)+(0.25, 0)$);%
\draw[myarrow] ($(Algo.south)+(0.25, 0)$) -- ($(Algo.south)+(0.25, -0.55)$) -- ($(Agent.south)+(-0.25, -0.55)$) node [below,pos=0.5] {reward $r_{t+1}$} -- ($(Agent.south)+(-0.25, 0)$);%
\draw[<->, thick, draw=black!32.5] (resBR.east |- resUL.center) -- ($(resBR.east |- resUL.center)+(0.375, -0.375)$);%
\path (resBR.east |- resUL.center)+(.325, -0.0975) node [text=black!32.5] {$\mathcal{C}$};%
\end{tikzpicture}%
\vskip-.75em
  \caption{Schematic of a cMDP with common state \& action spaces with transition, reward, and initial-state distributions conditioned on a context $c$. The diagram shows two learning pipelines: (i) \emph{context-oblivious} policy $\pi(s)$ receives only the state observation; (ii) \emph{context-aware} policy $\varpi(s,c)$ additionally observes the context. Both are trained across contexts $\context\sim\contextset$.}
  \label{fig:cMDP}\vskip-.75em
  \Description[Schematic of a Contextual MDP]{Schematic of a Contextual MDP (cMDP) with a common state-action space whose transition, reward, and initial-state distributions are conditioned on a context. The diagram shows two learning pipelines: (i) a context-oblivious policy -- pi of s -- that receives only the state observation s, and (ii) a context-aware policy -- varpi of s and c -- that additionally observes the context c. Both policies are trained across contexts sampled from a distribution.}
\end{figure}
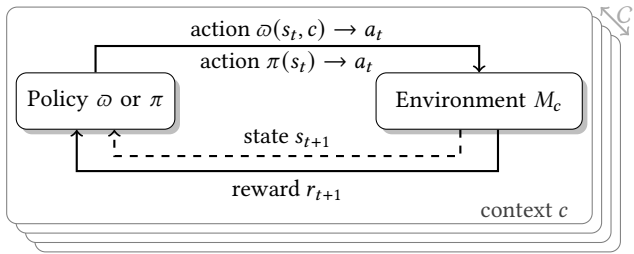

\emph{Contextual Markov Decision Processes (\CMDP)}~\citep{hallak-arxiv15a,modi-alt18a} (see \Cref{fig:cMDP}) extend the classic MDP formalism \citep{bellman-jmm57} to capture task generalization.
An MDP $\MDP = (\statespace, \actionspace, \transdomain, \rewards, \rho) $ comprises a state space $\statespace$, an action space $\actionspace$, transition dynamics $\transdomain$, a reward function $\rewards$ and a distribution over the initial states $\rho$.
By introducing a context variable $\context\in\contextset$, we can \emph{define}, \emph{characterize}, and \emph{parameterize} the environment's rules, thereby generating distinct task instances as contextual variations.
In a \CMDP, the action $\actionspace$ and state spaces $\statespace$ stay the same whereas the transition dynamics $\transdomain_\context$, rewards $\rewards_\context$ and initial state distributions $\rho_\context$ vary depending on the context $\context \in \contextset$.
Consequently, the context-dependent initial distribution and altered dynamics can expose the agent to different regions of the state space across contexts.
Following \citet{benjamins-tmlr23a} we allow the context space $\contextset$ to be discrete or described by a distribution $\contextdist$.
Thus, a \CMDP $\cMDP$ represents a family of related of MDPs $\cMDP = \{\MDPc \}_{\context\sim\contextdist}$ and can be seen as a sub-class of partially observable MDPs (POMDPs) \citep{ghosh-neurips21a}.

When learning policies $\pi$ on a \CMDP, one can choose to either train context-oblivious policies $\pi\colon\statespace\to\actionspace$ or context-aware ones $\varpi\colon\statespace\times\contextset\to\actionspace$. With domain randomization (DR) or procedural context generation (PCG), the goal is to learn policies that are robust to perturbations. Thus, learning agents typically can not explicitly\footnote{Relevant context might still be implicitly observable to an agent.}
access the context during learning, leading to context-oblivious policy $\pi$ that are robust to perturbations caused by sampling a new context $\context\sim\contextdist$. To facilitate learning with either DR or PCG, it is important to carefully choose the distribution $\contextdist$ to avoid exposing learning agents to highly different and potentially opposing experiences. Since the state $\statespace$ and action spaces $\actionspace$ are shared in a \CMDP, the same action might cause diametrically opposing outcomes (either in the reward function, the transition dynamics or both) for the same observed state. For example, assume a binary context and a binary action. Let the reward be the xor of the action and context values $\reward=a\oplus\context$ for a given state $\stateRL$. Since an agent does not know the context $\context$ in this example, it is impossible to figure out which action value is the optimal choice. While this toy example exaggerates the problem, it exemplifies why the choice of context distribution is highly critical for DR and PCG and often might need some form of curriculum learning approach to provide the most stable results \citep[see, e.g.][]{andrychowicz-ijrr20a,klink-neurips20a}. With dedicated curricula it is possible to control which experiences an agent is exposed to.

Context-aware policies $\varpi$ on the other hand can explicitly take context information into account when choosing an action in a particular state. Thus, such policies can avoid the issue presented in the previous paragraph since the context allows agents to differentiate between outcomes. However, a new complication now arises from the design decision on how to adequately incorporate context into learning policies. Commonly, context is treated as another observable element and thus is stacked to the (observation-)state vector $\varpi(\stateRL,\context) \to a$ \citep[see, e.g., ][]{perez2020generalized,biedenkapp-icml21a,sodhani-icml21a,zhang-iclr21a,yarats-icml21a,abdolshah-icml21a}. While this can help in learning more general policies it is no silver bullet \citep{benjamins-tmlr23a,ndir-ewrl24a}. Different algorithms seem to be more suited to this na\"ive treatment of context \citep{benjamins-tmlr23a} and this integration of context seems to not-trivially affect the learning dynamics, requiring potentially vastly different hyperparameter settings during training \citep{eimer-ecorl21a}.

A different line of work explores the use of hypernetworks \citep{hypernetworks} for contextual RL, in which such neural architectures learn to produce the weights of (or parts of) the policy networks, thereby adapting the policy's behavior to the context at hand \citep{beukman-neurips23a,benad2025shared,engwegen2025modular}. Counter to the simple concatenation, this approach enables a more dedicated context feature learning as context and observations are processed in two separate representation learning streams, before being merged in downstream layers. Importantly though, counter to approaches that simply learn context specific representations before merging them with observation specific features \citep{biedenkapp-prl22a,sparc-aaai26a}, hypernetwork approaches directly modify the policies behavior and do not simply condition policies on richer representations.

Opposite to the approaches that treat context largely as another observable, \citet{prasanna-rlc24a} try to exploit contextual information by more directly injecting context into latent representations of the Dreamer architecture \citep{dreamerv3-nature}. Thereby, instead of learning how different streams of knowledge interact, this injection can be seen as a form of modulation of the latent representations that have been learned from the observations. This approach enabled learning of policies that had better out-of-distribution generalization abilities, when compared with classic concatenation an domain randomization approaches. Crucially, the analysis highlighted that contextual knowledge allowed the world model to be more robust to counterfactual observations.
Similarly, \citet{gumbsch2024learning} aimed to learn when shifts in context occur (e.g., a door is opened/closed) to enable more proactive planning. Ultimately, this approach allowed them to learn temporal abstractions via hierarchies induced by context.

Furthermore, context-aware policies do not necessarily need explicit access to context. Instead, with system identification style approaches \citep{yu-rss17a,zhou-iclr19z,evans-icra22a}, as most often found in robotics \citep{iannotta2025contextbridgerealitygap} and meta-RL \citep{beck-tmlr25a}, attempt to estimate or recognize environment dynamics from a history of observations \citep{ndir-ewrl24a}. An important consequence of learning from a history of observations is that it enables online adaptation of context. Thus context is not treated as an unchanged quantity as it is done in the prior approaches. Essentially, context is quantified for a short snapshot and not for a whole episode or even longer period. \citet{ndir-ewrl24a} for example used this fact to learn context representations that are directly tailored to the current behavior of a policy, i.e., factors that are relevant to the states a policy will traverse through, rather than ones that globally aim to estimate a contexts influence on the transition dynamics. While this work was limited to small scale simulation settings, this approach has been recently demonstrated to enable policies to generalize even on real-world robotic hardware \citep{iannotta2025contextbridgerealitygap}.

Having formalized the learning of generalizable policies via cMDPs, we have presented two broad policy families that tackle this problem. For both we discussed lines of work that aim to address challenges of learning general policies. 
Across these lines of work a common thread emerges: \emph{the way context is represented, injected, and learned dramatically shapes the resulting learning dynamics}. While recent cRL methods have achieved impressive performance, they still treat context as a static, homogeneous signal. This limitation motivates a more nuanced view of context in which we recognize its heterogeneous nature and its evolution on multiple time-scales.  

In the next section we therefore propose a new taxonomy of contexts that classifies contexts by their structural properties, influence on dynamics, and temporal granularity. By making these distinctions explicit, we can design curricula, architectures, and learning mechanisms that fully exploit the rich, multi(-time)-scale character of context, moving us closer to genuine ``contextual intelligence''. 

\section{A Taxonomy of Context}\label{sec:tax}
Existing approaches treat the context variable as a monolithic, static signal. \citet{hallak-arxiv15a} first formalized cMDPs only considering static contexts, i.e., a context $c$ that is sampled once and remains unchanged while interacting with the MDP $\MDPc$ it instantiates. This is fundamentally misaligned with how intelligence operates in the real world where contexts can differ dramatically in \emph{what} they influence (dynamics, rewards, observations), \emph{how} they are presented to the agent (explicitly observable vs. latent), and \emph{when} they change (once per episode, intermittently within an episode, or continuously). To capture this heterogeneity we propose a taxonomy to decompose contexts into \emph{allogenic} (environment-imposed) and \emph{autogenic} (agent-driven) factors. A summary is provided in the \hyperlink{appendixlink}{appendix}.

\paragraph{Allogenic Context} Such contexts are \emph{exogenous}: they are imposed by the environment and are independent of the agent’s own actions. An agent can observe or infer them, but it cannot influence their evolution. As such, this form of context provides more global knowledge about the environments reward and transition dynamics. This form of context thus aligns with the notion of context introduced by \citet{hallak-arxiv15a}.
Some examples of allogenic contexts include
\begin{inparaenum}
    \item physical constants (gravity \citep{benjamins-tmlr23a}, length of limbs of a robot \citep{prasanna-rlc24a}, payload weights or atmospheric pressures);
    \item hardware variations (motor torque, sensor noise levels, center of mass \citep{9294027} or actuator latency);
    \item environment layouts (map topology , wall placement \citep{cobbe-icml20a}, terrain type or lighting conditions).
\end{inparaenum}

\paragraph{Autogenic Context} This type of contexts are \emph{endogenous}: they arise from the agent’s own behavior, internal state, or learning process and can therefore be influenced and even deliberately controlled by the policy. Consequently, this form of context is further removed from the notion of context as introduced in \citep{hallak-arxiv15a}. Some examples of these more agent-driven factors include
\begin{inparaenum}
  \item internal states (battery level, wear-and-tear of actuators, limb failure \citep{eimer-icml21a},  fatigue, or a hidden skill repertoire);
  \item self-generated task parameters (interaction frequencies \citep{biedenkapp-icml21a}, goals set by a higher-level planner \citep{NEURIPS2019_5c48ff18}, curriculum difficulty chosen by the agent, or the current sub-task).
\end{inparaenum}

Whereas allogenic contexts describe factors that the environment imposes on the agent and the thereby resulting behavior, autogenic contexts describe how agent behavior can shape the environment or the interaction with it.
This observation highlights a fundamental open problem: \textbf{how can we enable agents to reason jointly about allogenic and autogenic factors that exhibit fundamentally different influences and dependencies?} Existing cRL methods assume monolithic context and thus are not setup to exploit the heterogeneous structure revealed by our taxonomy. Addressing this gap requires new algorithmic primitives that 
\begin{inparaenum}
\item identify which aspects of the current situation are allogenic versus autogenic, 
\item condition policies appropriately on each type, and 
\item dynamically blend the two streams of information during learning and execution together with the regular observations.
\end{inparaenum}

\section{Temporal Hierarchy of Context}\label{sec:tempo}
While leveraging heterogeneous sources of contextual information already brings us closer to contextual intelligence, we believe that we can further exploit the structure of context by paying additional attention to the temporal nature of context.

Allogenic context represents global information about the transition dynamics. In most episodic settings these factors are approximately stationary: they remain essentially constant throughout an episode, exhibit only minor stochastic fluctuations, and only rarely undergo abrupt, large-scale shifts \citep[see, e.g.,][]{gumbsch2024learning}. For instance, consider a robot that must leave a paved footpath to let a pedestrian pass. The friction and compliance of the grass beside the path differ dramatically from those of the pavement; however, once the robot has switched surfaces, the relevant properties of the new surface stay roughly unchanged until another transition occurs.

Autogenic context however evolves as a direct consequence of the agents own actions and internal state. 
Such variables change more smoothly and more frequently within an episode, yet their evolution is still slower than that of raw observations. A concrete example is a robot’s battery charge: as the charge depletes, the robot preferentially selects low-energy maneuvers, causing the battery level to drift gradually rather than jump abruptly.

Allogenic contexts are largely piecewise-stationary and may experience sudden jumps, whereas autogenic contexts vary continuously and at a finer time-scale. Recognising and modelling these distinct temporal signatures is crucial for building agents that can both anticipate broad environmental changes and adapt fluidly to their own evolving internal state.
This observation highlights our second fundamental open problem: \textbf{how can we learn with sources of information that evolve at various different frequencies?} The typical cRL assumption of single monolithic context again will likely prohibit many approaches to directly exploit the notion of temporal dynamics discussed here. Notable exceptions here are the work by \citet{gumbsch2024learning} and the work on clockwork VAEs \citep{clockvaes} which have not yet been explored for cRL. To close this gap we need new algorithmic primitives that
\begin{inparaenum}
  \item exploit the temporal dynamics of allogenic and autogenic contexts (e.g., via multi-timescale representation learning \citep{clockvaes} or change-point detection);
  \item balance exploration and exploitation with respect to both context streams (probing the environment to detect abrupt allogenic shifts versus exploiting the current autogenic context \citep{griesbach2025learning});
  \item integrate the two streams with the raw observation stream, adhering to the temporal structures without drowning out dynamic information with static ones (e.g., via dedicated encoder branches whose representations are dynamically blended during execution).
\end{inparaenum}

\section{Beyond Physical Quantities}
In recent developments, the focus of cRL has predominantly been directed towards physical quantities. While this is particularly enticing with the outlook on embodied AI, we believe that cRL research should broaden its focus to include more abstract context types \citep[e.g.,][]{rajan-jair23a,mohan-jair24a,bordne-automlws24a}.
Our taxonomy admits \emph{abstract} contexts that might not be directly measurable but are crucial for many MAS applications.

\paragraph{Team Roles}
In various multi-agent reinforcement learning (MARL) settings \citep{albrecht2024multi}, agents may adopt different functional roles within an environment.
This is particularly evident in cooperative settings, where a team of agents might have distinct roles that must be filled to achieve a common goal. In a soccer team, for example, one can distinguish between defenders, midfielders, and attackers. While all agents share the same objective, they must exhibit different behaviors to ensure success.
These roles, however, need not remain static throughout a game. If, for example, a defender possesses the ball in front of an empty goal, it should recognize that, momentarily, offensive behavior is required beyond what its normal role would permit.
Thus, via dedicated communication protocols, roles might be exchanged when agents fulfill the necessary requirements.

Furthermore, in MARL settings, some agents might be human and thus require different coordination strategies than artificial agents. When approaching a human agent during a package delivery scenario, an artificial agent should prioritize the human's safety. However, when approaching another artificial agent in the same scenario, these safety considerations might not apply, allowing for a different approach procedure.

\paragraph{Resource Awareness}
Previous examples have already elaborated on how battery power could inform more appropriate decision making in robotic agents. However, resource awareness can also comprise allogenic components. An autonomous factory or robotics warehouse could exploit knowledge of resource availability to increase production during peak renewable energy generation and reduce it to essential operations when only non-renewable energy is available. This extends beyond energy considerations and allows a manufacturing system to adjust its production schedule based on material availability or even carbon credit budgets.

Similarly, a more simple gardening robot might reduce water usage during droughts or defer watering if the weather forecast reliably predicts rain. In agricultural settings, such systems could further consider soil moisture levels, seasonal water allocation permits, and competing demands from other agricultural zones, demonstrating how resource contexts can be both physical and regulatory in nature.
Meanwhile, energy, compute, and network bandwidth are naturally modeled as continuous autogenic variables. By conditioning policies on such contexts, robots can automatically trade off task urgency against resource consumption wich is particularly essential for long-duration missions.

\paragraph{Regulatory / Ethical Context}
Finally, our world is governed not only by the laws of physics, but also by those of our societies \citep{ec2019ethics}.
A self-driving car might leverage the fact that the German Autobahn has no general speed limit, but must immediately adhere to speed limits upon crossing into neighboring countries. Such agents must further adapt to right-of-way rules, permitted lane-changing behaviors, and even culturally-specific expectations about pedestrian interactions. What is considered polite yielding behavior in one country might be seen as dangerous hesitation in another~\citep{russel-book}.

Legal regimes (such as speed zones) and ethical constraints (such as privacy budgets) and other forms of human preferences are allogenic yet abstract.
Adequately encoding this form of context will be far from trivial but necessary to enable safe coexistence in our shared physical world. The challenge lies not just in representing these constraints, but in enabling agents to reason about their interactions. Thus, while RL from human feedback \citep{christiano-neurips17a} has helped in shaping preference based-reward signals, we do not believe that it is enough to indirectly expose agents to human preferences but explicitly condition their behavior on these preferences.

These examples illustrate that contextual intelligence extends beyond low-level physics; it encompasses any factor that influences optimal decision-making. This observation highlights our final fundamental open problem: \textbf{how can we incorporate high-level and abstract contexts into the learning process?} To tackle this challenge an important aspect is to expose non-physical contexts to learning agents in settings where they might already be available or otherwise derive simulators that enable us to make progress towards this open problem. 

\section{Conclusion}
\balance
The future of autonomous agents hinges on contextual intelligence. 
While the field has made great progress, often in isolated domains, we argue that crucial characteristics of what constitute context have not yet been taken into account. A unified view will result in agents that understand the difference between what they can change and what they must adapt to, that operate across temporal scales, and that reason about abstract contexts like roles and regulations will enable entirely new classes of applications.
We call on the community to abandon the notion that context is merely another input feature. Context is the foundation on which we can build dedicated architectures, learning mechanisms, and theoretical frameworks. Otherwise, RL will remain confined to narrow domains with perfect simulators.

\begin{acks}
André Biedenkapp acknowledges funding through the research network “Responsive and Scalable Learning for Robots Assisting Humans” (ReScaLe) of the University of Freiburg. The ReScaLe project is funded by the Carl Zeiss Foundation.
\end{acks}

\bibliographystyle{ACM-Reference-Format} 
\bibliography{bib/strings,bib/local,bib/lib,bib/proc}

\onecolumn
\appendix\hypertarget{appendixlink}{}
\section*{Appendix}
\paragraph{On the Relativity of Context}
We ultimately believe that context is relative. The frame of reference determines if a context is allogenic, autogenic, or somewhere in between and requires careful consideration when setting up the learning pipeline.
In a MARL setting, e.g., this depends on whether you view the context from an individual agent’s frame of reference or a central critic's. One frame might see a context as allogenic (a global, static quantity), while another sees it as autogenic (local and dynamic).

More concretely, in the MARL setting, the presence of other agents could be viewed as either allogenic or autogenic and depends on how agents interact.
If other agents follow a plan regardless of the learning agent’s choices (e.g., "blind" execution), this provides allogenic context. Knowing this is valuable because the learning agent can treat them as a stable, physical constraint to avoid or utilize.
If other agents react to the learning agent's choices, the context is autogenic.
Note that this further emphasizes that allogenic context being more constant does not mean that the environment can not change. Rather, allogenic context has a globally the same effect in the environment and rarely changes. 

\paragraph{Taxonomy Summary}
\begin{table}[h!]
    \caption{Taxonomy Summary Table}
    \label{tab:taxonomy_summary}
    \centering
    \begin{tabularx}{\textwidth}{
    >{\raggedright\arraybackslash\hsize=.5\hsize}X
    >{\raggedright\arraybackslash\hsize=.5\hsize}X 
    >{\raggedright\arraybackslash\hsize=.9\hsize}X
    >{\raggedright\arraybackslash\hsize=.85\hsize}X
    >{\raggedright\arraybackslash\hsize=.85\hsize}X
    >{\raggedright\arraybackslash\hsize=2.0\hsize}X
    }
    \toprule
        \textbf{Taxonomy Level} & \textbf{Influence} & \textbf{Source} & \textbf{Temporal Nature} & \textbf{Type of Change} & \textbf{Examples} \\
    \midrule
        Allogenic & Global & 
        Exogeneous, Imposed by the environment, Independent of Agent's actions & 
        Approximately stationary, Infrequent change & 
        Abrupt \citep[see, e.g., ][]{gumbsch2024learning}, Large scale & 
        Physical constants (gravity \citep{benjamins-tmlr23a}, payload weights, \ldots); hardware variations (motor torque, center of mass \citep{9294027}, \ldots); environment layouts (map topology, wall placement \citep{cobbe-icml20a}, lighting conditions, \ldots) \\
    \midrule
        Autogenic & Local & 
        Endogeneous, Arise from agent's own behavior, internal state, or learning process & 
        Frequent change, Slower change than observations & 
        More smoothly, Continuous & 
        Internal states (battery level, limb failure \citep{eimer-icml21a}, fatigue, \ldots); Self-generated task parameters (interaction frequencies \citep{biedenkapp-icml21a}, goals set by a higher-level planner \citep{NEURIPS2019_5c48ff18}, curriculum difficulty, \ldots) \\
    \bottomrule
    \end{tabularx}
\end{table}
\Cref{tab:taxonomy_summary} provides a summary overview of the Taxonomy as introduced in \Cref{sec:tax} and the corresponding temporal natures of \Cref{sec:tempo}.

\end{document}